\documentclass[11pt]{article}

\usepackage{acl}

\usepackage{times}
\usepackage{latexsym}

\usepackage[T1]{fontenc}

\usepackage[utf8]{inputenc}

\usepackage{microtype}

\usepackage{color}
\usepackage{import}
\usepackage{xstring}
\usepackage[export]{adjustbox}
\usepackage{booktabs}
\usepackage{caption}
\usepackage{chronology}
\usepackage{enumitem}
\usepackage{float}
\usepackage{hyperref}
\usepackage{makecell}
\usepackage{multirow}
\usepackage{subfigure}
\usepackage{tabu}
\usepackage{tabularx}
\usepackage{pifont}
\usepackage{colortbl}
\usepackage{csquotes}
\usepackage{tikz}
\usepackage{pifont}
\usepackage{boldline}

\title{ROAST: Review-level Opinion Aspect Sentiment Target Joint Detection for ABSA}

\author{Siva Uday Sampreeth Chebolu\textsuperscript{1}, Franck Dernoncourt\textsuperscript{2}, \\
{\bf Nedim Lipka\textsuperscript{3}},  
{\bf Thamar Solorio\textsuperscript{1,4}}  \\
        \textsuperscript{1}University of Houston, 
        4800 Calhoun Rd, Houston, TX, 77004, USA \\ 
        \textsuperscript{2}Adobe Research, 
        801 N 34th St, Seattle, WA 98103, USA  \\
        \textsuperscript{3}Adobe Research, 
        321 Park Ave, San Jose, CA, 95110, USA  \\
         \textsuperscript{4}MBZUAI, 
        Masdar City, Abu Dhabi, UAE  \\
        \textsuperscript{1}\texttt{\{sivauday.sampreeth8,thamar.solorio\}@gmail.com} \\ \textsuperscript{2}\texttt{franck.dernoncourt@gmail.com}, 
        \textsuperscript{3}\texttt{lipka@adobe.com}
        }

\begin{document}
\maketitle
\begin{abstract}

Aspect-Based Sentiment Analysis (ABSA) has experienced tremendous expansion and diversity due to various shared tasks spanning several languages and fields and organized via SemEval workshops and Germeval. Nonetheless, a few shortcomings still need to be addressed, such as the lack of low-resource language evaluations and the emphasis on sentence-level analysis. To thoroughly assess ABSA techniques in the context of complete reviews, this research presents a novel task, Review-Level Opinion Aspect Sentiment Target (ROAST). ROAST seeks to close the gap between sentence-level and text-level ABSA by identifying every ABSA constituent at the review level. We extend the available datasets to enable ROAST, addressing the drawbacks noted in previous research by incorporating low-resource languages, numerous languages, and a variety of topics. Through this effort, ABSA research will be able to cover more ground and get a deeper comprehension of the task and its practical application in a variety of languages and domains (\url{https://github.com/RiTUAL-UH/ROAST-ABSA}).
\end{abstract}

\section{Introduction}
\label{Sec: Introduction}
The ABSA field has gained significant attention, leading to multiple shared tasks organized by SemEval workshops, including SemEval-2014 Task 4, SemEval-2015 Task 12, and SemEval-2016 Task 5 \cite{14-dataset,15-dataset,16-dataset}. The 2014 SemEval ABSA (SE-ABSA14) focused on English reviews, annotating aspect words, their polarity for laptop and restaurant reviews, and aspect categories and their polarity for restaurants. The 2015 SemEval ABSA (SE-ABSA15) introduced a new framework, connecting identified opinion elements like aspects, opinion target expressions, and sentiment polarities within tuples. The aspect category combines an entity type E and an attribute type A, distinguishing between entities and the aspects being assessed. SE-ABSA16 expanded the task to other languages and domains, such as Chinese, Turkish, and Spanish reviews.

Germeval hosted an ABSA task for the German language reviews in 2017, with some differences from SE16-ABSA \cite{germevaltask2017}. SemEval-2022's structured sentiment analysis task \cite{22-dataset} is similar to ABSA but combines aspect-based sentiment, targeted sentiment, end2end, and sentiment target extraction and does not include aspect categories. 
These advancements have significantly enriched the ABSA landscape, bringing forth unique nuances in annotation and expanding into various languages and domains. However, these developments have also exposed certain limitations that require further exploration.

One of the prominent limitations pertain to the ASQP, ACOS, and OATS dataset \cite{ASQP,ACOS,OATS-ACL}. While OATS spans multiple domains, including relatively unexplored areas such as Amazon Fine Foods and Coursera course reviews, it exclusively comprises English-language reviews. This constraint is notable, considering the extensive diversity of languages in which user-generated content, including reviews, is generated. The absence of reviews in low-resource languages is a limitation we acknowledge, and we are committed to addressing it in future endeavors. As highlighted in our ABSA Datasets survey, the availability of datasets in low-resource languages is limited, constraining the development and applicability of ABSA models in these contexts.

While OATS represents a significant step forward in addressing some of these limitations by providing valuable insights into the intricacies of ABSA it does not fully encompass the real-world challenges that ABSA models may encounter, such as the need for evaluation at different levels of ABSA. 

    \begin{figure*}[!ht]
        \centering
        \includegraphics[width=0.8\textwidth]{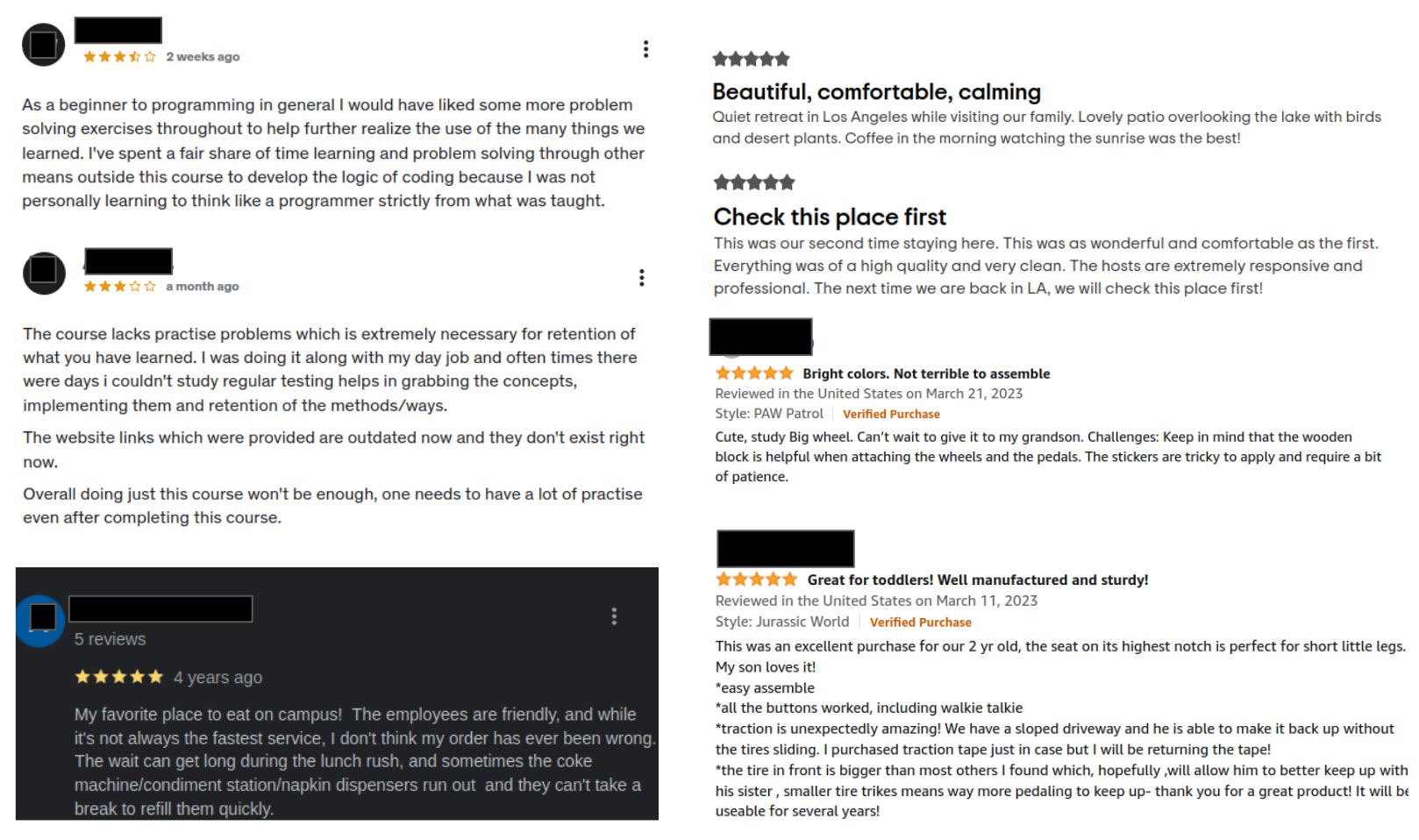}
        \caption{Multi sentence reviews from different platforms.}
        \label{fig: Multi-sent-reviews}
    \end{figure*}

In real-world examples, such as Figure \ref{fig: Multi-sent-reviews}, reviews often contain multiple sentences with overlapping context dependency. Sentence-level ABSA methods fail to generalize to full reviews, as they assign a \textit{NULL} value for implicit targets and opinions. However, considering the entire review context, a target might be explicitly mentioned and referred to implicitly in other sentences. Previous ABSA tasks focused on review sentences rather than entire reviews, limiting their applicability to real-world situations. Although SE16-ABSA included a Text-Level ABSA subtask, it aimed to summarize opinions from individual review sentences instead of capturing all four ABSA elements within the entire review context.

We aim to address this with the introduction of a novel task: ROAST (Review-Level Opinion Aspect Sentiment Target). ROAST emphasizes the comprehensive evaluation of ABSA methods at the review level. It responds to the call to bridge the gap between sentence-level ABSA, which is limited in its ability to capture the nuances of complete reviews, and text-level ABSA, which often lacks the necessary granularity. ROAST seeks to fill this void by introducing the concept of joint detection of all ABSA elements within the context of full reviews.

ROAST introduces the concept of joint detection of all ABSA elements at the review level, offering a holistic perspective on reviews' sentiments, aspects, targets, and opinions. The datasets we create for the ROAST task address the limitations we found in OATS by extending the coverage to multiple languages, including low-resource ones, and by venturing into new domains. This expansion is particularly crucial given the insights from our ABSA Datasets survey, which emphasized the need for more challenging datasets to reflect real-world scenarios spanning diverse domains and languages.

In essence, while OATS lays the foundation for understanding ABSA elements and advancing joint modeling techniques, ROAST extends this foundation to encompass the broader evaluation of ABSA methods in real-world scenarios, multiple languages, and diverse domains. Together, these initiatives aim to drive ABSA research toward a better understanding of the task and its real-time applicability, addressing the diverse challenges posed by reviews across different languages and domains.



\section{Related Works}
\label{Sec: Related Works}
In this section, we describe the tasks of ABSA and their related datasets, followed by triplet and quadruple extraction task-specific methods.

\subsection{Current Datasets}
Traditionally, when ABSA was first introduced by \citet{Hu_2004} as a task in NLP, the main aim then was to extract different aspects given a sentence followed by assigning polarities given those aspects. 
As a decade passed, \citet{14-dataset} added another important element of ABSA: aspect categories. In the SemEval-2014 shared task on ABSA, two sub-tasks were drafted in addition to the aspect term extraction and its sentiment polarity. They are the aspect category detection and their sentiment polarity. Two datasets from the restaurant and laptop domains were released as part of this shared task. 

In the consecutive years, \citet{15-dataset,16-dataset} proposed two more ABSA shared tasks at the SemEval, redefining the subtasks of ABSA and giving them a more concrete structure. 
Aspect categories were a combination of entity and attribute pairs, aspect terms were called targets, and the sentiment polarity was assigned jointly for the targets and the aspect categories. 
Several datasets with the three elements from multiple languages and domains were published, including English, Dutch, Spanish, French, Russian, and Turkish restaurants, English laptops, Arabic hotels, and many others. 

Later, \citet{TargetAspectSentimentJD} utilized the data from the SemEval-2015 and SemEval-2016 to define the target-aspect-sentiment joint detection task (TASD). After that, researchers felt the importance of detecting opinion phrases to identify the sentiment polarity and understand its relationship with targets and aspect categories. 
This resulted in three new sub-tasks with new datasets that stemmed from the SemEval challenges, namely aspect-opinion-pair extraction (AOPE) \cite{AOPE}, target-opinion-word extraction (TOWE) \cite{TOWE}, and aspect-sentiment-triplet-extraction (ASTE) \cite{ASTE-dataset}. 
More recently, the ABSA research has shifted towards the joint detection of all four elements of ABSA, which proved to better identify the inter-relationships among the elements, thereby enhancing the performance of other subtasks. 
This task is called aspect sentiment quadruple prediction (ASQP) \cite{ASQP} or aspect-category-opinion-sentiment joint detection (ACOS) \cite{ACOS}. The ASQP task is evaluated on the dataset with quadruples created from the SemEval challenges, ASTE triplets, and TOWE and AOPE tuples. 
The ACOS task introduced two datasets from the restaurant and the laptop domain with implicit and explicit aspect terms and opinion phrases. 

The current ABSA dataset landscape is notably skewed towards reviews and particular domains such as restaurants and electronics. This stems from the easy availability and volume of data on review websites and other online platforms. 
This narrow focus restricts the wider application of ABSA, limiting its ability to deliver insights across different sectors. 
Expanding the diversity of ABSA datasets is essential to push research and applications beyond the boundaries of mere reviews. To address this limitation, we propose three new datasets for the quadruple extraction task from three new domains, along with implicit and explicit targets and opinion phrase annotations. We also include the review-level tuples for identifying the overall sentiment polarity for different aspect categories in a review. This dataset could be used to solve all the above-mentioned subtasks of ABSA. 

\subsection{Related Methods}
We focus on a few joint detection tasks: the TASD, ASTE, and ASQP for our research. \citet{sampreeth-ABSA-Datasets-Survey-arxiv} have provided a detailed survey of the other datasets, tasks, and their challenges. 

\textbf{Triplet Extraction Tasks} 
In ASTE research, three primary methodologies have emerged: MRC-based techniques \cite{BMRC}, methods anchored on BERT and table-filling \cite{BDTF}, and generative approaches \cite{gen-scl-nat,GAS-T5,ASQP}. The MRC-based methods involve crafting a specific query for each component in the triplet, subsequently extracting them based on the response to this query. Generative methods, in contrast, frame the ASTE challenge as a sequence generation task and employ sequence-to-sequence (seq2seq) models. The triplets are then decoded using a specially tailored algorithm. In this study, we employ representative techniques from these three categories and investigate OATS using these methods. 

For the TASD task, there are two paths: BERT-based methods \cite{TargetAspectSentimentJD}, and generative-based approaches \cite{T5-ABSA-sampreeth,ASQP}. The BERT-based methods extract the aspect sentiment tuple from a sentence using the sentence-pair classification as a backbone \cite{14-Utilizing-Bert} followed by extracting the targets for each pair using the token classification with the BIO or a softmax classifier. The generative-based approaches convert the task similar to an abstractive summarization and use the sequence-to-sequence models like the T5, BART, and others \cite{T5,bart} to predict the triplets. 

\textbf{Quadruple Extraction Task}
Researchers have pointed out two promising directions. \citet{ACOS} propose a two-stage method by extracting the aspect and opinion terms first. Then, these items are utilized to classify aspect categories and sentiment polarity. Another method is based on the generation model \cite{ASQP}. By paraphrasing the input sentence, the quadruplet can be extracted end-to-end. In this work, we follow the generative direction and consider the order-free property of the quadruplet. A few other studies also proposed generative-based models with this paraphrasing as a backbone, including \citet{Template-Quad,gen-scl-nat}. 

\section{Task Formulation}
\label{Sec: Task Formulation}
In contrast to the Aspect Sentiment Quadruple Prediction (ASQP) task, which operates at the sentence level, the Review-level Opinion Aspect Sentiment Target (ROAST) task seeks to produce sentiment quadruples for an entire review. This approach aims to address the challenge of implicit targets and opinions that are contextually dependent on the entire review rather than individual sentences.

    Given an opinionated review R, comprising a series of sentences {s1, s2, ..., sn}, the ROAST task is to predict all review-level sentiment quadruples {(c, a, o, p)}, where each quadruple consists of an aspect category c from a predefined category set C; an aspect term a and an opinion term o, which may span across sentences or be implied within the review context, i.e., $a \in R_{xy} \cup \{\emptyset\}$ and $o \in R_{xy} \cup \{\emptyset\}$, with $R_{xy}$ denoting the set containing all possible continuous or non-continuous text spans of R; and a sentiment polarity p, which falls into one of the sentiment classes \textit{positive, negative, neutral, and conflict}. This formulation captures the complexity of sentiment analysis when opinion expressions about a target may be dispersed throughout the review or inferred from the overall context.

\subsection{Challenges of ROAST}
    The transition from sentence-level ABSA, as applied to the OATS dataset, to review-level analysis with the ROAST dataset introduces several computational and linguistic challenges. Methods that efficiently and effectively processed the OATS dataset may face the following challenges when adapted to the ROAST dataset. 
    
    \paragraph{Contextual Complexity} Unlike ASQP, ROAST must handle the increased complexity of sentiment expressed over multiple sentences, which requires understanding the flow of sentiment across the review and resolving references to targets and opinions mentioned earlier or later in the text.

    \paragraph{Computational Intensity} The extended length of reviews implies a higher computational demand for processing and analyzing the data. This can significantly impact the resource requirements for both training and inference phases, potentially limiting the scalability of the ROAST task. 
    The increase in sequence length necessitates a corresponding increase in the token limits for models like T5 and BERT, leading to larger model sizes and reduced batch sizes to fit within the confined computational space. 
    Moreover, the generative models struggle with the auxiliary format in which they are fine-tuned when confronted with the generation of numerous opinions from the longer reviews. The inherent limitations in token generation capacity lead to not capturing potentially insightful opinions, resulting in incomplete sentiment analysis.

    \paragraph{Co-reference Resolution} The task involves resolving co-references where targets and opinions may be referred to with pronouns or other indirect expressions, necessitating advanced co-reference resolution techniques that can work effectively over long text spans. 
    For instance, a review might state, "The camera quality is breathtaking, and it makes the phone worth every penny. However, the processor performance of it is not that good." Here, the first occurrence of "it" refers to the "camera quality," which is an aspect that needs to be connected to the positive sentiment expressed later in the sentence. In contrast, the second one refers to the phone itself. 
    In the context of the ROAST task, the challenge is amplified by the extended length of reviews, which often include multiple sentences with numerous anaphoric links that weave through the narrative of the user's experience. 


    \paragraph{Aspect Category Disambiguation} Although ROAST is expected to help reduce the aspect category misclassifications that arise from the implicit targets by using the additional context, disambiguating aspect categories in a review is still challenging. 
    The reason may be that a review speaks about multiple categories simultaneously, often blurring the lines between them. 
    The methods must accurately attribute sentiments to the correct aspect category, which may be subtly indicated within the text. 

    \paragraph{Linguistic Challenges}  Languages such as Hindi and Telugu present unique challenges due to subtle variations in letters that significantly alter meanings and complicate the extraction of correct opinion phrases. The task of linking these phrases with their respective targets is further complicated by the increased review length, which disperses relevant contextual cues and opinion expressions across a wider span of text. 

\section{Dataset}
\label{Sec: Dataset}
All the past ABSA datasets are at sentence level and irrelevant for evaluating the ROAST task. So, we curated 6 new datasets from 5 distinct domains in three languages to evaluate this task. 
    We will discuss about the sources, annotations procedure, and several statistics of the dataset in this section. 
    Table \ref{tab: reviews-quads} (in appendix) presents a high-level overview of the datasets with sources, example reviews, and their quadruples for all languages and domains. 

    \paragraph{Amazon\_FF, Coursera, and Hotels (English)} The data for these 3 domains is an extension to the OATS datasets \cite{OATS-ACL}. We added 266, 300, and 560 additional reviews from the same sources, respectively. 
    
    \paragraph{Phones (English and Hindi)} We crawled several Indian review websites for phone reviews and cleaned the data for HTML tags, code-switching instances, and other unwanted text before giving it to the annotators. We ended up with only 618 reviews that were available in Hindi directly. So, we took an additional 1464 English phone reviews and translated them manually into Hindi for this corpus. We followed the SE16-ABSA task for the phones dataset and used a similar list of aspect categories with some additions and removing a few unwanted categories. We also provide the quadruples and overall tuples for the 1464 English Phone reviews along with the Hindi reviews. 
    
    \paragraph{Movies Dataset (Telugu)} We collected the Telugu movie reviews by crawling the \url{www.123telugu.com}, where they have different sections discussing about the story, what is good and bad in the movie, technical aspects, and overall verdict or bottom line of the review. We took only the technical aspects and the bottom line or review summary for each movie and removed unwanted symbols, URLs, emojis, section titles, and other irrelevant text in the data cleaning step. 

    \subsection{Annotation Procedure}
        We hired two annotators through Upwork for the Amazon\_FF, Coursera, hotels, and phones domains and one annotator for the Telugu movie reviews to annotate each review with the opinion quadruples using the BRAT tool. 
        The annotation process is designed in two phases. 
        In phase one, both annotators are given 50 distinct reviews from all the domains to annotate for the quadruples following a guidelines document drafted for this task. 
        Then, each annotator reviewed the annotations the other annotator gave, and they resolved the disagreements collectively. 
        When annotators lacked assurance, a decision was taken in collaboration with me to arrive at an acceptable judgment. 
        Following that, each annotator is given 50 common reviews from each domain (except Telugu movies) to measure the inter-annotator agreement using the F1-score metric on the opinion quadruples. 
        
        For Phase Two, we divided the remaining reviews equally among the annotators to complete the annotation task. For every 50 reviews, the annotations given by one annotator for 10 reviews are reviewed by another to ensure the quality of the annotations. If there are any disagreements, that portion of 50 reviews is checked another time to fix similar issues.        
        We annotated approximately 2000 reviews for all domains except Telugu movies, which comprise 1000 reviews. 
        We spent \$11000 for annotating the datasets, which resulted in \$1.4 per review, which took an average time of 5 minutes for the annotator. 
        We identified the average F1-score across domains was nearly 78\%, which is very reasonable given the subjectivity of the task with two span-extraction elements (targets and opinion phrases) in the quadruple. We provide the detailed annotation scores in Table \ref{tab: ROAST-IAA}.

        \begin{table}[!ht]
            \caption{Inter-Annotator Agreement scores for the ROAST Datasets}
            \label{tab: ROAST-IAA}
            \centering
            \scriptsize
            \resizebox{0.9\columnwidth}{!}{%
            \begin{tabular}{lrrrr}
            \hline
            \textbf{Domain} &
              \multicolumn{1}{l}{\textbf{Tgt-Op. Ph.}} &
              \multicolumn{1}{l}{\textbf{AC-Pol}} &
              \multicolumn{1}{l}{\textbf{Quadruples}} &
              \multicolumn{1}{l}{\textbf{ASD-Tuples}} \\ \hline
            {Amazon\_FF}    & 85.77 & 94.69 & 82.81 & 93.72 \\
            {Coursera}      & 91.04          & 93.91          & 87.81          & 92.35          \\
            {Hotels}        & 86.26          & 90.47          & 81.64          & 89.84          \\
            {Phones\_Eng}   & 73.87          & 90.69          & 69.67          & 86.82          \\
            {Phones\_Hindi} & 72.11          & 89.27          & 69.53          & 82.98          \\ \hline
            \end{tabular}%
            }
            
        \end{table}


    \subsection{Statistics}
        In this section, we present and discuss several statistics of the ROAST datasets by looking from different perspectives of sentiment analysis. 
        First, we will look at the basic numbers for each dataset in Table \ref{tab: ROAST-Overall}, which has the total number of reviews, quadruples, and overall review-level category-polarity tuples. On the other hand, Table \ref{tab: ROAST-Avg} gives the aggregate statistics with the average number of quadruples and tuples per review and the maximum and average number of tokens per review.

        \begin{table}[!ht]
            \caption{ROAST Dataset overall statistics}
            \label{tab: ROAST-Overall}
            \centering
            \scriptsize
            \resizebox{0.8\columnwidth}{!}{%
            \begin{tabular}{lrrr}
            \hline
            \multicolumn{1}{l}{\textbf{Domain}} & \multicolumn{1}{l}{\textbf{\#Reviews}} & \multicolumn{1}{l}{\textbf{\#Quadruples}} & \multicolumn{1}{l}{\textbf{\#Tuples}} \\ \hline
            {Amazon\_FF}           & 2060           & 7681           & 4592           \\
            {Coursera}             & 2061           & 8833           & 6202           \\
            {Hotels}               & 2056           & 15258          & 10493          \\
            {Phones\_Eng}          & 1464           & 9473           & 6960           \\
            {Phones\_Hindi}        & 2083           & 13979          & 10114          \\
            {Movies}               & 1643           & 21631           & 16220           \\ \hline
            {Total}                & {11367} & {76855} & {54581} \\ \hline
            \end{tabular}%
            }
            
        \end{table}

        \begin{table}[!ht]
            \caption{ROAST Dataset aggregate statistics}
            \label{tab: ROAST-Avg}
            \centering
            \scriptsize
            \resizebox{\columnwidth}{!}{%
            \begin{tabular}{lrrrr}
            \hline
            \textbf{Domain} &
              \multicolumn{1}{l}{\textbf{Avg. \#Quads}} &
              \multicolumn{1}{l}{\textbf{Avg. \#Tuples}} &
              \multicolumn{1}{l}{\textbf{Max \#Tokens}} &
              \multicolumn{1}{l}{\textbf{Avg. \#Tokens}} \\ \hline
            {Amazon\_FF}           & {3.91} & {2.34} & {215} & {76.79}  \\
            {Coursera}             & {4.31} & {3.03} & {224} & {76.41}  \\
            {Hotels}               & {7.53} & {5.18} & {277} & {84.78}  \\
            {Phones\_Eng}          & {6.57} & {4.83} & {284} & {84.67}  \\
            {Phones\_Hindi}        & {6.97} & {5.04} & {384} & {102.47} \\
            {Movies}               & {13.16} & {10.84} & {301} & {115.87} \\ \hline
            \end{tabular}%
            }
            
        \end{table}

        We annotated one of the biggest datasets for the quadruple prediction task within the ABSA community, which has 10,304 reviews with 63,313 quadruples and 44,111 category-polarity tuples. 
        The hotels domain has the highest number of opinions among all the datasets, and Amazon\_FF with the least. 
        Amazon\_FF domain has the most personal statements rather than opinions on the product, resulting in the lowest quadruples to reviews ratio (3.91) among all the domains. Similarly, the Coursera domain has second to the worst ratio of number of quadruples to reviews (4.31). 
        In contrast, the movies domain has the highest quadruple-to-reviews ratio (14.5) and is way ahead of the other domains. 
        The average number of tokens per review also varies among the domains, with Coursera having the shortest reviews (nearly 77 tokens) and movies being the longest (nearly 116 tokens). 
        The average number of tuples indicates how well the opinions are distributed among several aspect categories. For instance, there are only 3 aspect categories on which the quadruple opinions are expressed. On the other hand, the movies domain has opinions on at least 10 distinct aspect categories per review on average. 

        Table \ref{tab: ROAST-Opinion} provides a comprehensive breakdown of the distribution of reviews within the ROAST Datasets based on the number of opinions in each review. As we go from left to right in the columns, the number of opinions per review increases, where the \#0-Op column gives the statistics of reviews with no opinions, and the rightmost column \#Multi-Op denotes the reviews with more than four opinions.         
        
        \begin{table}[!ht]
            \caption{ROAST Dataset Opinion statistics. \#\{x\}-Op: indicates number of reviews with \{x\} opinions. \#Multi-Op: Number of reviews with more than 4 opinions. }
            \label{tab: ROAST-Opinion}
            \centering
            \scriptsize
            \resizebox{\columnwidth}{!}{%
            \begin{tabular}{lrrrrrr}
            \hline
            \textbf{Domain} &
              \multicolumn{1}{l}{\textbf{\#0-Op}} &
              \multicolumn{1}{l}{\textbf{\#1-Op}} &
              \multicolumn{1}{l}{\textbf{\#2-Op}} &
              \multicolumn{1}{l}{\textbf{\#3-Op}} &
              \multicolumn{1}{l}{\textbf{\#4-Op}} &
              \multicolumn{1}{l}{\textbf{\#Multi-Op}} \\ \hline
            {Amazon\_FF}           & {96}   & {120} & {332} & {429} & 426  & 657  \\
            {Coursera}             & {11}   & {91}  & {272} & {405} & 418  & 864  \\
            {Hotels}               & {29}   & {12}  & {46}  & {77}  & 140  & 1752 \\
            {Phones\_Eng}          & {22}   & {9}   & {53}  & {109} & 178  & 1093 \\
            {Phones\_Hindi}        & {77}   & {25}  & {69}  & {151} & 234  & 1527 \\
            {Movies}               & {145} & {0}   & {1}   & {1}   & 5    & 1491  \\ \hline
            {Total}                & 235          & 258          & 772          & 1172         & 1398 & 6447 \\ \hline
            \end{tabular}%
            }
            
        \end{table}

        Notably, the number of reviews, as indicated in the last row of the table, increases progressively from left to right. This progression underscores the versatility and richness of the ROAST Datasets. Particularly striking is the finding that more than 87\% of the dataset's reviews encompass more than two distinct opinions. This highlights the dataset's depth and complexity, making it an ideal resource for ABSA research. The increasing prevalence of multi-opinion reviews further emphasizes the real-world relevance of the dataset, aligning with the multifaceted nature of opinions expressed in various domains. 
        
        One primary rationale behind proposing the ROAST task and creating this corpus is to reduce the number of implicit targets and opinions using the additional context available from the entire review rather than dealing with individual sentences independently. Moreover, getting explicit targets and opinions will help disambiguate the aspect categories and sentiment polarity tasks, further helping the quadruple extraction task. 
        Table \ref{tab: ROAST-Exp-Imp} provides the distribution of explicit and implicit targets and opinion phrases in the ROAST dataset. It reveals that explicit targets and explicit opinion phrases are predominant in most domains, with explicit opinions significantly outnumbering implicit opinions, supporting our thought process. 

        \begin{table}[!ht]
            \caption{Distribution of Explicit and Implicit Targets (ET and IT) and Explicit and Implicit Opinion Phrases (EO and IO) in the ROAST Dataset with their different combinations. }
            \label{tab: ROAST-Exp-Imp}
            \centering
            \scriptsize
            \resizebox{\columnwidth}{!}{%
            \begin{tabular}{lllrrrr}
            \hline
            \textbf{Domain} &
              \textbf{ET / IT} &
              \textbf{EO / IO} &
              \multicolumn{1}{l}{\textbf{ET-EO}} &
              \multicolumn{1}{l}{\textbf{ET-IO}} &
              \multicolumn{1}{l}{\textbf{IT-EO}} &
              \multicolumn{1}{l}{\textbf{IT-IO}} \\ \hline
            {Amazon\_FF}    & 5,864 / 1,811 & 7,399 / 276  & 5,628  & 236 & 1,771 & 40 \\
            {Coursera}      & 8,205 / 631   & 8,389 / 447  & 7,769  & 436 & 620   & 11 \\
            {Hotels}        & 14,569 / 691  & 14,816 / 444 & 14,139 & 430 & 677   & 14 \\
            {Phones\_Eng}   & 8,883 / 605   & 9,439 / 49   & 8,836  & 47  & 603   & 2  \\
            {Phones\_Hindi} & 13,140 / 839  & 13,911 / 68  & 13,072 & 68  & 839   & 0  \\
            {Movies}        & 21,606 / 25     & 21,387 / 244   & 21,362   & 244  & 25     & 0  \\ \hline
            \end{tabular}%
            }
            
        \end{table}

\section{Experiments}
\label{Sec: Experiments}
We chose the three most significant joint detection tasks for our review-level ABSA, two triplet extraction tasks, and one quadruple extraction task. In the triplet extraction, we chose target-aspect-sentiment detection (TASD) \citep{TargetAspectSentimentJD} and aspect sentiment triplet extraction (ASTE) \citep{ASTE-dataset}. We use the proposed ROAST task for the quadruple prediction, which is like a review-level aspect sentiment quadruple prediction (ASQP) \citep{ASQP}. Again, we chose these joint detection tasks with a rationale that joint models that are tailored to manage multiple intertwined elements simultaneously have consistently been shown to surpass models fine-tuned for single-element extraction in various domains \citep{DREAM-paper,T5-ABSA-sampreeth}. 

\subsection{Baseline Methods}
    \label{SubSec: Baseline-Methods}
        Similar to the \citet{OATS-ACL} experiments, we implement several representative models from various frameworks, including MRC-based, generation-based, and BERT-based frameworks, for review-level task evaluations. 
    
        \paragraph{Task-specific methods} 
    
            For \textit{ASTE}, we use the following methods:
            
            \textbf{BMRC} \cite{BMRC}: a MRC-based method. It extracts aspect-oriented triplets and opinion-oriented triplets. Then, it obtains the final results by merging the two directions. 
    
            \textbf{GEN-NAT-SCL} \cite{gen-scl-nat}: a generative-based method. It combines a new generative format with a supervised contrastive learning objective to predict the ASTE triplets and quadruples. 
    
            \textbf{BDTF (BERT)} \cite{BDTF}: a BERT-based method that uses table-filling to solve the problem. It transforms the ASTE task into detecting and classifying relation regions in the 2D table representing each triplet in addition to an effective relation representation learning approach to understand word and relation interactions. 
    
            
    
    
            \paragraph{TASD Task-specific methods} 
            For the TASD task, we chose \textbf{TAS-BERT} \cite{TargetAspectSentimentJD}, which is a BERT-based method. It fine-tunes the pre-trained BERT model to solve the aspect-sentiment detection task using the classification token and then detects the targets corresponding to those tuples using the token classification with CRF/softmax decoding. 
        
        \paragraph{Unified methods} 
            The following generative frameworks can be applied to any sentence-level and review-level ABSA tasks:
    
            \textbf{GEN-NAT-SCL} \cite{gen-scl-nat}: a generative-based method. It combines a new generative format with a supervised contrastive learning objective to predict the ASTE's triplets and ASQP's quadruples. 
            
            \textbf{GAS} \cite{GAS-T5}: a generation-based method. It transforms the ASTE, ASQP, TASD, and ASD tasks into a text generation problem that inputs the review and generates all the respective combinations of ABSA opinion elements as output. 
            
            \textbf{Paraphrase} \cite{ASQP}: a generative-based method. It is similar to GAS but transforms output opinion elements into paraphrases that read as natural sentences. We substituted the corresponding element for implicit targets and opinions with the word "it."


\subsection{Evaluation}
\label{SubSec: Evaluation}
    Following \citet{OATS-ACL,ASQP} experiments, we use the F1 score to measure the performance of different approaches on all the tasks. All experimental results are reported using the average of 5 different runs using distinct random seeds. We divided each domain dataset into train, validation, and test sets with 70\%, 10\%, and 20\% splits, respectively. A triplet and quadruple is considered correct only if all the corresponding prediction elements match the gold standard labels. We consider any partial matches as wrong predictions following \citet{ASQP}. We adopt the base versions of all the transformer models for our experiments on all the English language datasets, including the BERT-base \citep{BERT}, T5-base \citep{T5}, RoBERTa-base \citep{roberta}, and others. For the GEN-SCL-NAT method, we used T5-large, the default setting for that approach. 
    For the Hindi and Telugu languages, we adopted multilingual pretrained transformer models, especially MT5-base, for the GAS and Paraphrase methods. 
    We finetuned all the tasks using the 23GiB Tesla P40 GPUs with a batch size 8 for Amazon\_FF and Coursera domains, 4 for the hotels domain, and 2 for the phones and movies domains. We tuned the max\_sequence\_length hyper-parameter ranging from 128 to 512, based on the dataset's average and median number of tokens.

\section{Results and Discussion}
\label{Sec: Results and Discussion}
In this section, we will present the results of several experiments we performed using the above-mentioned baseline methods on all the ROAST datasets.  
    First, we will look at the triplet extraction task results, review-level TASD and ASTE task results, followed by the quadruple extraction task, ROAST. 

    Table \ref{tab: ROAST-ASTE} shows the performance of the ASTE task on all the English language ROAST datasets. 
    The BMRC and BDTF (BERT) methods exhibit mixed results, with some experiments terminating due to GPU memory constraints (OOM) in the Hotels and Phones domains. 
    The input sequence length of these methods should match the increased context length of the input reviews, and on average, the number of opinion quadruples per review is higher for the hotels and the phones domains, resulting in the out of GPU memory errors for these domains. 
    Besides this downside, BMRC is the best performing method in the Coursera domain. 
    The GAS and Paraphrase methods manage to maintain competitive performance levels across all domains. 
    Amazon\_FF domain is the most challenging domain with the best performance at 35.49 F1-score using the Paraphrase method. Coursera and hotels domains are not so far from each other. 
    
    \begin{table}[!ht]
        \centering
        \caption{Results for Review-level ASTE task on \underline{ROAST English} Datasets. OOM: experiments ran out of GPU memory. }
        \label{tab: ROAST-ASTE}
        \resizebox{0.85\columnwidth}{!}{%
        \begin{tabular}{@{}lrrrr@{}}
        \toprule
        \multicolumn{1}{l}{\textbf{Method}} &
          \multicolumn{1}{c}{\textbf{Amazon\_FF}} &
          \multicolumn{1}{c}{\textbf{Coursera}} &
          \multicolumn{1}{c}{\textbf{Hotels}} &
          \multicolumn{1}{c}{\textbf{Phones}} \\ \midrule
        BMRC        & 32.18 & \textbf{51.41} & OOM   & OOM   \\
        BDTF (BERT) & 18.96 & 41.62 & OOM   & OOM   \\ 
        GAS         & 33.46 & 50.20 & \textbf{51.75} & 37.69 \\
        Paraphrase  & \textbf{35.49} & 49.76 & 51.41 & \textbf{43.94} \\ \bottomrule
        \end{tabular}%
        }
    \end{table}

    Table \ref{tab: ROAST-TASD} presents the results for the Review-level TASD task conducted on the ROAST English Datasets. 
    The TAS-BERT-BIO and TAS-BERT-TO methods exhibit competitive performance in the TASD task. TAS-BERT-TO, in particular, demonstrates a notable increase in F1 scores in the Coursera domain compared to Amazon\_FF, showcasing its adaptability to different domains. 
    The GAS and Paraphrase methods consistently outperform the TAS-BERT variants across all domains in terms of F1 scores. 
    Across all domains, Paraphrase consistently performs slightly better than GAS in the TASD task. For example, in the Amazon\_FF domain, Paraphrase achieves an F1 score of 47.57, compared to GAS's F1 score of 44.38. 
    Hotel reviews seem the easiest domain with a 61.84\% F1-score, and Amazon\_FF is, as always, the most challenging one. 
    Of the two triplet extraction tasks, the TASD task is easier across all domains. This is because the ASTE task poses a challenge in identifying the opinion phrases.

    \begin{table}[!ht]
        \centering
        \caption{Results for Review-level TASD task on \underline{ROAST English} Datasets }
        \label{tab: ROAST-TASD}
        \resizebox{0.85\columnwidth}{!}{%
        \begin{tabular}{@{}lrrrr@{}}
        \toprule
        \multicolumn{1}{l}{\textbf{Method}} &
          \multicolumn{1}{c}{\textbf{Amazon\_FF}} &
          \multicolumn{1}{c}{\textbf{Coursera}} &
          \multicolumn{1}{c}{\textbf{Hotels}} &
          \multicolumn{1}{c}{\textbf{Phones}} \\ \midrule
        TAS-BERT-BIO & 42.11 & 47.72 & 48.49 & 43.87 \\
        TAS-BERT-TO  & 42.74 & 55.87 & 43.93 & 41.64                \\
        GAS          & 44.38 & 54.45 & 60.49 & 49.31                \\
        Paraphrase   & \textbf{47.57} & \textbf{55.44} & \textbf{61.84} & \textbf{53.56}                \\ \bottomrule
        \end{tabular}%
        }
    \end{table}

    We present the ROAST task experimental results in Table \ref{tab: ROAST-ASQP}. 
    Again, we observed that the Paraphrase method outperformed the GAS and GEN-SCL-NAT methods on all the domains. Specifically, on the phone reviews, using complete sentences in terms of paraphrases helped the model better identify the quadruples. Given that the performance on the TASD task is significantly higher than the ROAST, it is evident that identifying the opinion phrases and linking them to the correct target is the most challenging task. It again reconfirms the lower scores of the same methods on the ASTE task compared to the TASD task. 
    
    \begin{table}[!ht]
        \centering
        \caption{Results for the ROAST task on \underline{ROAST English} Datasets }
        \label{tab: ROAST-ASQP}
        \resizebox{0.85\columnwidth}{!}{%
        \begin{tabular}{@{}lrrrr@{}}
        \toprule
        \multicolumn{1}{l}{\textbf{Method}} &
          \multicolumn{1}{c}{\textbf{Amazon\_FF}} &
          \multicolumn{1}{c}{\textbf{Coursera}} &
          \multicolumn{1}{c}{\textbf{Hotels}} &
          \multicolumn{1}{c}{\textbf{Phones}} \\ \midrule
        GEN-NAT-SCL & 24.02 & 35.44 & 43.77 & 33.65 \\
        GAS         & 23.56 & 35.41 & 44.62 & 32.09 \\ 
        Paraphrase  & \textbf{26.11} & \textbf{37.27} & \textbf{46.03} & \textbf{39.28} \\ \bottomrule
        \end{tabular}%
        }
    \end{table}

    Table \ref{tab: ROAST-Non-Eng} gives the performance scores on all three tasks for the non-English datasets, Hindi phones, and Telugu movies. GAS and Paraphrase have alternatively good performances across the datasets. 
    For the Hindi phone reviews, both methods identified the triplets almost equally on the ASTE task. However, the Paraphrase method outperformed the GAS method significantly on the TASD and the ROAST tasks, with F1-scores of 51.52\% and 35.83\%, respectively. 
    On the other hand, GAS performed better on the Telugu movie reviews for the ASTE and ROAST tasks, with 52.57 and 40.95, respectively, while Paraphrase was slightly better on the movie reviews TASD task. 

    \begin{table}[!ht]
        \centering
        \caption{Results for \underline{Hindi Phones} and \underline{Telugu Movies} ROAST Datasets}
        \label{tab: ROAST-Non-Eng}
        \resizebox{0.9\columnwidth}{!}{%
        \begin{tabular}{@{}lrrrrrr@{}}
        \toprule
        \multirow{2}{*}{\textbf{Method}} &
          \multicolumn{2}{c}{\textbf{ASTE}} &
          \multicolumn{2}{c}{\textbf{TASD}} &
          \multicolumn{2}{c}{\textbf{ROAST}} \\ \cmidrule(l){2-7} 
         & \textbf{Phones} & \textbf{Movies} & \textbf{Phones} & \textbf{Movies} & \multicolumn{1}{l}{\textbf{Phones}} & \multicolumn{1}{l}{\textbf{Movies}} \\ \midrule
        GAS        & \multicolumn{1}{r}{32.62} & \multicolumn{1}{r}{\textbf{52.57}} & \multicolumn{1}{r}{45.36} & \multicolumn{1}{r}{40.84} & 29.94 & \textbf{40.95} \\
        Paraphrase & \multicolumn{1}{r}{\textbf{32.97}} & \multicolumn{1}{r}{42.03} & \multicolumn{1}{r}{\textbf{51.52}} & \multicolumn{1}{r}{\textbf{43.68}} & \textbf{35.83} & 33.15 \\
        GEN-SCL-NAT             & -            &  -          &  -           & -           & 35.55                      & 36.06                      \\ \bottomrule
        \end{tabular}%
        }
    \end{table}

    \paragraph{ROAST (vs) OATS}

        In Table \ref{tab: ROAST-OATS}, we present a comparative analysis of explicit (ET) and implicit (IT) targets, as well as explicit opinions (EO) and implicit opinions (IO), between the OATS and ROAST datasets. This comparison sheds light on the distribution of opinion-related elements and the evolution from OATS to ROAST, highlighting important trends and insights.
        
        \begin{table}[!ht]
            \caption{Comparison of OATS (vs) ROAST for Explicit and Implicit Opinion Distribution}
            \label{tab: ROAST-OATS}
            \centering
            \scriptsize
            \resizebox{\columnwidth}{!}{%
            \begin{tabular}{lllll}
            \toprule
            \multirow{2}{*}{\textbf{Domain}} &
              \multicolumn{2}{c}{\textbf{Target (ET / IT)}} &
              \multicolumn{2}{c}{\textbf{Opinion Phrase (EO / IO)}} \\ \cmidrule(l){2-5} 
             &
              \multicolumn{1}{c}{\textbf{OATS}} &
              \multicolumn{1}{c}{\textbf{ROAST}} &
              \multicolumn{1}{c}{\textbf{OATS}} &
              \multicolumn{1}{c}{\textbf{ROAST}} \\ \hline
            {Amazon\_FF} & 2,999 / 5,261 & 5,864 / 1,811 & 6,780 / 1,480  & 7,399 / 276  \\
            {Coursera}   & 5,163 / 2,712 & 8,205 / 631   & 6,185 / 1,690  & 8,389 / 447  \\
            {Hotels}     & 8,654 / 2,820 & 14,569 / 691  & 10,193 / 1,281 & 14,816 / 444 \\ \bottomrule
            \end{tabular}%
            }
            
        \end{table}

        Firstly, we observed a big shift in the ratio of explicit to implicit targets in all the domains, especially Amazon\_FF, which changed from dominant implicit targets to explicit targets (0.57 to 3.06). Hotels gained the highest difference of this ratio, increasing from 3.23 to 21.08. 
        Similarly, the opinion phrases ratio also got a positive drift in all the domains. For instance, the Amazon\_FF moved from 4.58 to 26.81 and hotels from 7.95 to 33.36. 
        These trends indicate that, taking the context of the entire review, we could mitigate the implicit opinions to a maximum extent, along with staying close to real-world scenarios. 

        In the Coursera and hotels domain, ROAST exhibits around 28.0\% and approximately 45.0\% more explicit opinion phrases than OATS.  In the Amazon\_FF and hotels domains, ROAST features approximately 96.9\% and 68.2\% more explicit targets than OATS. 

        In Table \ref{tab: ROAST-Run-Time}, we present a comparative analysis of the experiment run-times per epoch (in minutes) for various BERT-based and Generative approaches applied to the ROAST datasets. The top section focuses on BERT-based methods, while the bottom section showcases generative approaches employing T5. These run-time metrics offer insights into the efficiency and computational demands of different techniques across diverse datasets.

        \begin{table}[!ht]
            \centering
            \caption{Comparison of experiment run-time per epoch (in mins) for several BERT-based and Generative approaches. The top rows are all BERT-based approaches, and the bottom ones are generative approaches using T5. OOM indicates that the experiments terminated with out of GPU memory errors. (-): indicates that the method is not used for that dataset}
            \label{tab: ROAST-Run-Time}
            \resizebox{\columnwidth}{!}{%
            \begin{tabular}{@{}lrrrrrr@{}}
            \toprule
            \textbf{Method} &
              \multicolumn{1}{c}{\textbf{Amazon\_FF}} &
              \multicolumn{1}{c}{\textbf{Coursera}} &
              \multicolumn{1}{c}{\textbf{Hotels}} &
              \multicolumn{1}{c}{\textbf{Phones\_Eng}} &
              \multicolumn{1}{c}{\textbf{Phones Hindi}} &
              \multicolumn{1}{c}{\textbf{Movies Telugu}} \\ \midrule
            BMRC        & 25  & 32  & OOM & OOM & -                     & -                     \\
            BDTF (BERT) & 35  & 45  & OOM & OOM & -                     & -                     \\
            TAS-BERT    & 360 & 435 & 521                     & 510                     & -                      & -                      \\ \midrule
            GEN-NAT-SCL & 12  & 13  & 18                      & 13                      & \multicolumn{1}{r}{8} & \multicolumn{1}{c}{2} \\
            GAS         & 1   & 1   & 2                       & 1                       & \multicolumn{1}{r}{2} & \multicolumn{1}{c}{2} \\
            Parphrase   & 1   & 1   & 2                       & 1                       & \multicolumn{1}{r}{2} & \multicolumn{1}{c}{2} \\ \bottomrule
            \end{tabular}%
            }
        \end{table}

        For BERT-based methods, we observe that BMRC and BDTF (BERT) exhibit similar run-time patterns across domains such as Amazon\_FF and Coursera. However, for more extensive datasets like Hotels, Phones\_Eng, and Phones\_Hindi, both methods face resource constraints, leading to out-of-memory (OOM) errors. TAS-BERT, a more complex BERT-based approach, necessitates significantly longer run-times, particularly evident in the Hotels and Phones\_Eng domains, emphasizing the computational demands of this method. We couldn't apply the TAS-BERT to the Telugu movies and Hindi phones domains because they have bigger sequences compared to the English datasets, and this would result in quadratic time and excessive computational requirements to complete the experiments.

        In contrast, the generative approaches, GEN-NAT-SCL and GAS, demonstrate substantially shorter run-times across all domains, even when handling larger datasets like Hotels and Phones\_Eng. These methods efficiently generate results while maintaining manageable computational overhead. The Paraphrase approach follows a similar trend, showcasing efficient run times across various domains.
        
        The notable aspect to highlight is the divergence in computational efficiency between BERT-based and generative approaches. While BERT-based methods are more resource-intensive, generative techniques like GEN-NAT-SCL, GAS, and Paraphrase deliver results with considerably shorter run times. These findings underscore the trade-off between model complexity and computational efficiency, providing valuable insights for researchers when selecting appropriate methods for NLP tasks on the ROAST datasets.


\section{Conclusions}
\label{Sec: Conclusions}
In conclusion, this work provides a comprehensive exploration of the ROAST task, encompassing the creation of diverse datasets, their statistical overview, meticulous experimental design, and insightful comparison analysis between the OATS and ROAST datasets. The ROAST task underscores the complexity of ABSA in the context of review-level text data, offering valuable insights into opinion phrase extraction and target identification across various domains. Notably, the findings reveal that sentence-level ABSA methods may not be ideally suited for the ROAST task, primarily due to the requirement to consider the entire review as context. This necessitates adjustments for GPU memory constraints and the handling of larger input sequences with decent batch sizes, which are critical aspects to address in future research.

Beyond the challenges observed, this work highlights several key takeaways. Generative approaches such as GAS and Paraphrase methods emerge as robust performers, demonstrating their versatility and promise in tackling the multifaceted ABSA tasks presented in the ROAST datasets. 
Additionally, distinct domain-specific variations in task difficulty underscore the importance of domain adaptation strategies when deploying ABSA models in practical applications.

Looking ahead, future endeavors in the realm of ROAST task research hold immense potential. Cross-domain experiments, cross-lingual investigations, and the exploration of cross-domain cross-lingual ABSA represent compelling avenues for further study. These endeavors could provide a deeper understanding of the transferability of sentiment analysis models across domains and languages, enhancing the applicability of ABSA techniques in real-world scenarios. 

\section{Limitations and Ethics}
\label{Sec: Limitations}
We point out the following ethical considerations while building and using the ROAST dataset:
\begin{enumerate}
    \item The data collected from platforms such as Amazon Finefoods, Coursera, TripAdvisor Hotels, Amazon Phones, and Telugu movie review platforms come from public reviews provided by users. We ensured that all identifiable information, including usernames, avatars, and any other potentially identifying details, were removed to preserve the anonymity of the reviewers.
    \item Our data processing methodology focused on extracting and analyzing the content of the reviews without altering the original sentiment or meaning. We handled this data with the utmost care to ensure that the sentiments and opinions of the original reviewers were not misrepresented.
    \item We recognize that reviews from online platforms might not represent the complete spectrum of users' opinions, as they may be influenced by various factors like platform algorithms, user demographics, and more. We urge users of this dataset to be aware of potential biases and always consider the data in the proper context.
    \item While the reviews were publicly available, the original authors might not have expected their content to be used in research. We've taken measures to respect their privacy, but future users of this dataset should also be aware of this consideration.
    \item While the reviews are publicly accessible, we acknowledge the platforms (Amazon, Coursera, TripAdvisor, and 123telugu) as the source of this data. We have only used this data for research and academic purposes, ensuring that we respect the terms of use for each platform.
\end{enumerate}


\bibliography{main}
\bibliographystyle{acl_natbib}



\end{document}